\documentclass[10pt,twocolumn,letterpaper]{article}

\usepackage[pagenumbers]{wacv} 

\usepackage{graphicx}
\usepackage{amsmath}
\usepackage{amssymb}
\usepackage{booktabs}
\usepackage{times}
\usepackage{epsfig}
\usepackage{graphicx}
\usepackage{amsmath}
\usepackage{amssymb}
\usepackage{booktabs}
\usepackage{algorithm}
\usepackage{algorithmic}
\usepackage{adjustbox}
\usepackage{array}
\usepackage{multirow}
\usepackage{float}
\usepackage{enumitem}
\usepackage[accsupp]{axessibility}
\usepackage[table]{xcolor}
\definecolor{light-gray}{gray}{0.9}
\newcommand{\mycc}{\cellcolor{light-gray}}

\usepackage{etoolbox}
\makeatletter
\patchcmd{\maketitle}
 {\def\@makefnmark}
 {\def\@makefnmark{}\def\useless@macro}
 {}{}
\makeatother

\usepackage[pagebackref,breaklinks,colorlinks]{hyperref}

\usepackage[capitalize]{cleveref}
\crefname{section}{Sec.}{Secs.}
\Crefname{section}{Section}{Sections}
\Crefname{table}{Table}{Tables}
\crefname{table}{Tab.}{Tabs.}

\def\wacvPaperID{803} 
\def\confName{WACV}
\def\confYear{2024}

\begin{document}

\title{Effective Restoration of Source Knowledge in Continual Test Time Adaptation}




\author{Fahim Faisal Niloy$^{1}$, Sk Miraj Ahmed$^{1}$, Dripta S. Raychaudhuri$^{2,\ast}$,\thanks{Currently at AWS AI Labs. Work done while the author was at UCR.} Samet Oymak$^{3}$, Amit K. Roy-Chowdhury$^{1}$\\
$^{1}$University of California, Riverside ~$^{2}$AWS AI Labs ~$^{3}$University of Michigan, Ann Arbor\\
{\tt \small \{fnilo001@, sahme047@, drayc001@, amitrc@ece.\}ucr.edu, oymak@umich.edu}}

\maketitle

\begin{abstract}
 Traditional test-time adaptation (TTA) methods face significant challenges in adapting to dynamic environments characterized by continuously changing long-term target distributions. These challenges primarily stem from two factors: catastrophic forgetting of previously learned valuable source knowledge and gradual error accumulation caused by miscalibrated pseudo labels. To address these issues, this paper introduces an unsupervised domain change detection method that is capable of identifying domain shifts in dynamic environments and subsequently resets the model parameters to the original source pre-trained values. By restoring the knowledge from the source, it effectively corrects the negative consequences arising from the gradual deterioration of model parameters caused by ongoing shifts in the domain. Our method involves progressive estimation of global batch-norm statistics specific to each domain, while keeping track of changes in the statistics triggered by domain shifts. Importantly, our method is agnostic to the specific adaptation technique employed and thus, can be incorporated to existing TTA methods to enhance their performance in dynamic environments. We perform extensive experiments on benchmark datasets to demonstrate the superior performance of our method compared to state-of-the-art adaptation methods.
\end{abstract}

\section{Introduction}
Deep neural networks (DNNs) have demonstrated remarkable success in numerous applications. However, they are known to suffer from performance degradation when faced with distributional shifts between the training and test data. This poses a significant risk in deploying DNNs in domains such as autonomous driving or  medical imaging, where encountering unseen types of test data during deployment could result in undesirable consequences. To overcome this challenge, test-time adaptation (TTA) \cite{wang2020tent, niu2022efficient} has emerged as a promising approach. 

TTA aims to adapt DNNs to the unseen target domain using only unlabeled test data streams, without the need for a substantial portion of the test data to be available as in traditional domain adaptation settings~\cite{wang2020tent}, and without accessing the source data used for training the model. This enables efficient and on-the-fly adaptation to distribution shifts, making it computationally efficient and highly effective in real-world scenarios.

\begin{figure}[t!]
\centering
\includegraphics[width=0.5\textwidth]{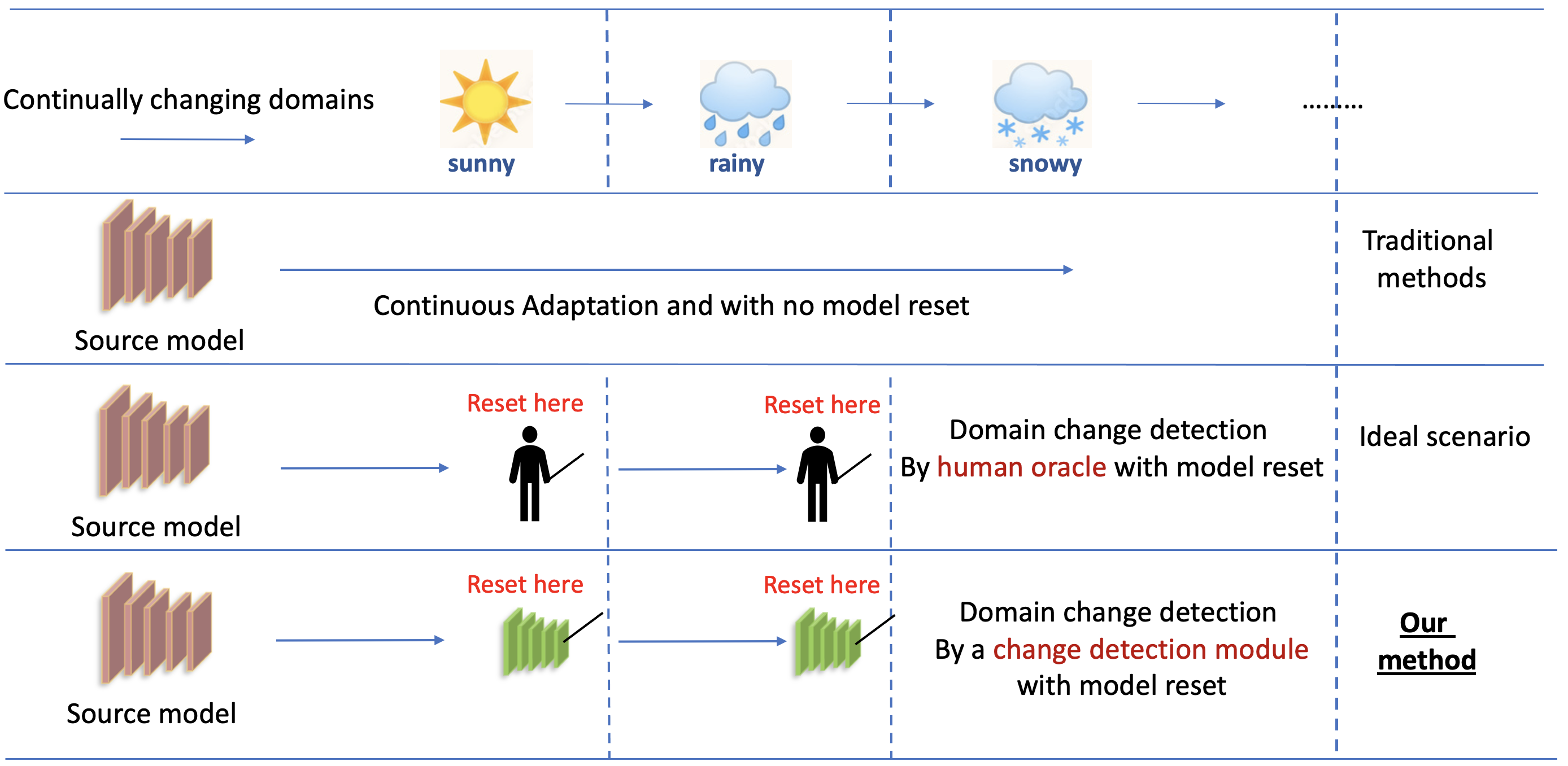} 
\caption{\textbf{Problem setup.} Traditional test-time adaptation methods suffer from forgetting and error accumulation over time as they continuously adapt to incoming target distributions. Previous research has demonstrated the benefits of resetting the model to its original parameters when a domain change occurs, but these approaches rely on an oracle with additional domain knowledge for detecting such changes. In this paper, we propose an automated strategy to detect domain changes, allowing for efficient and effective model resets in practical scenarios.}


\vspace{-6pt}
\label{fig:teaser}
\end{figure}

Existing TTA approaches typically focus on adapting to distribution shifts between a fixed source domain and a target domain. However, in real-world scenarios, the target domain may not remain static and can continuously evolve. For instance, a self-driving car model trained on data from clear weather conditions may encounter test data from diverse weather conditions such as rain, snow, fog, etc. In such cases, it is crucial for the model to adapt its parameters to these new domains accordingly in a continual fashion to ensure optimal performance. This particular setting is often referred to as continual test time adaptation in the literature ~\cite{wang2022continual, song2023ecotta}. Current TTA methods mostly fail to account for such dynamic domain changes due to two primary reasons: 

\begin{itemize}[leftmargin=*,topsep=0pt]
\setlength\itemsep{-2pt}
\item \textbf{Catastrophic Forgetting}: Over extended periods of continuous adaptation to new distributions, the model may experience catastrophic forgetting, wherein the knowledge learned from the source domain is gradually lost~\cite{wang2022continual}. This can potentially erase valuable information learned from the source domain and can have a negative impact on the model's performance when adapting to subsequent new domains.
\item \textbf{Error Accumulation}: Several TTA methods \cite{wang2020tent, zhang2022memo} leverage pseudo-labels for unsupervised adaptation. In a continually changing environment, the dynamic distribution shift can cause the pseudo-labels to become progressively noisier and miscalibrated. Thus, early prediction errors are more prone to propagate and accumulate over time, potentially leading to error accumulation.
\end{itemize}



To effectively tackle the challenges of catastrophic forgetting and error accumulation in dynamic environments, we propose to \emph{detect the underlying domain changes} and \emph{restore the source knowledge accordingly}. Specifically, we observe that the KL divergence between pre-trained batch-norm statistics and the batch-norm statistics of incoming test batches can effectively quantify domain shifts. Based on this insight, we develop an approach to detect domain changes in dynamic environments. We maintain an exponential running average of the batch-norm statistics of incoming test batches to progressively estimate the global batch-norm statistics specific to each domain, using an adaptive momentum value determined by the KL divergence between the running batch-norm statistics and the incoming test batch statistics. When a sudden distribution shift occurs in the incoming test batches, the momentum value undergoes a sharp change. By monitoring this, we can identify outliers caused by domain shifts. 

Using this approach, our method detects changes in the distribution of a dynamic environment and appropriately restores the model's source knowledge. The primary objective is to prevent the forgetting of previously acquired source knowledge and address the cumulative negative impact of noisy and miscalibrated pseudo-labels. As a result, our method enables more effective model adaptation. This makes it highly suitable for dynamic scenarios where the target domain may experience shifts over time.

Also, our domain change detection module is independent of the adaptation method used. Therefore, our proposed module can be seamlessly integrated into existing TTA methods, enhancing their robustness to distribution changes in dynamic environments. It is important to note that the test time adaptation literature \cite{wang2022continual, niu2022efficient} often assumes an `online' version of the models, where an oracle with additional ground-truth domain knowledge is available to restore the source knowledge when a domain change occurs. Such online models are inherently resilient to the forgetting issue. However, the requirement of ground-truth knowledge renders the models impractical for application in real-world scenarios. This paper is the first attempt to develop an approach that effectively serves as this oracle, yet without requiring any ground-truth domain knowledge, enabling the practical implementation of such online models.

\noindent \textbf{Main contributions.} To summarize, our primary contributions are as follows: 

\begin{itemize}[leftmargin=*,topsep=0pt]
\setlength\itemsep{-2pt}
\item We address a novel problem of detecting domain changes in the context of continual test time adaptation, which focuses on dynamic environments where the target distribution undergoes continual domain shifts. 

\item We demonstrate that detecting such domain changes and thereby restoring the source knowledge has the ability to mitigate catastrophic forgetting and error accumulation. 

\item We extensively evaluate our proposed method on real-world datasets, encompassing a wide range of tasks. Through these experiments, we provide substantial empirical evidence that demonstrates the effectiveness and applicability of our approach.
\end{itemize}

\section{Related Work}

\noindent\textbf{Unsupervised Domain Adaptation.} Unsupervised domain adaptation (UDA) aims to enhance the performance of a pre-trained source model when there is a distribution shift between the labeled source domain and the unlabeled target domain. UDA has been extensively applied in various computer vision tasks, including image classification \cite{tzeng2017adversarial}, semantic segmentation \cite{tsai2018learning}, object detection \cite{hsu2020progressive}, and reinforcement learning \cite{raychaudhuri2021cross}. Existing approaches typically focus on aligning the distributions of the source and target domains using techniques like maximum mean discrepancy \cite{long2015learning}, adversarial learning \cite{ganin2016domain, tzeng2017adversarial}, and more. Recently, there has been a growing interest in adaptation using only a pre-trained model without the need for source data. This shift is motivated by privacy and memory storage concerns associated with the source data. These approaches leverage techniques such as information maximization \cite{liang2020we, ahmed2021unsupervised, ahmed2022cross}, pseudo-labeling \cite{yeh2021sofa, kumar2023conmix}, and self-supervision \cite{xia2021adaptive}.

\noindent \textbf{Test Time Adaptation.} 
 UDA methods typically require a significant amount of target domain data to adapt a model, regardless of whether they utilize source data. Thus, the adaptation process is performed offline, meaning that it happens before the model is deployed or used for inference on the target domain. In contrast, test time adaptation (TTA) approaches adapt a model to the target data after deployment, i.e., during inference or testing phase. It involves updating the model's parameters or internal representations during inference based on the characteristics of the current test batch from the target domain to improve performance on the subsequent test batches. TTA approaches also do not require the source data to be available during adaptation.
 
 In an early work \cite{li2016revisiting}, authors leverage the batch-norm statistics of incoming test batches to adapt the model to the target distribution instead of relying on pre-trained batch-norm statistics. TENT \cite{wang2020tent} adapts a pre-trained source model on incoming target data by minimizing entropy and updating the batch-norm parameters of the source model. DUA \cite{mirza2022norm} continuously updates the batch-norm statistics of the pre-trained source model with the incoming test batches in order to align to the target distribution. TTA methods have also been extended to the segmentation task \cite{valanarasu2022fly, shin2022mm, liu2021source, hu2021fully}.

TTA methods can also be used to adapt during inference to a dynamically varying target distribution, that is, where the target distribution changes after different time intervals. In this case, TTA methods usually suffer from the problem of error accumulation and catastrophic forgetting - continually drifting away from source knowledge. Few approaches have been proposed to address these issues. CoTTA \cite{wang2022continual} applies stochastic restoration of source knowledge such that the model does not drift much away from source knowledge. EATA \cite{niu2022efficient} introduces a regularization loss to ensure that important model weights are preserved during adaptation, thereby alleviating forgetting. In contrast to these methods, our approach provides a more structured and effective solution to mitigate forgetting and error accumulation, resulting in better performance.


\section{Method}

\begin{figure}[t!]
\centering
\includegraphics[width=0.4\textwidth]{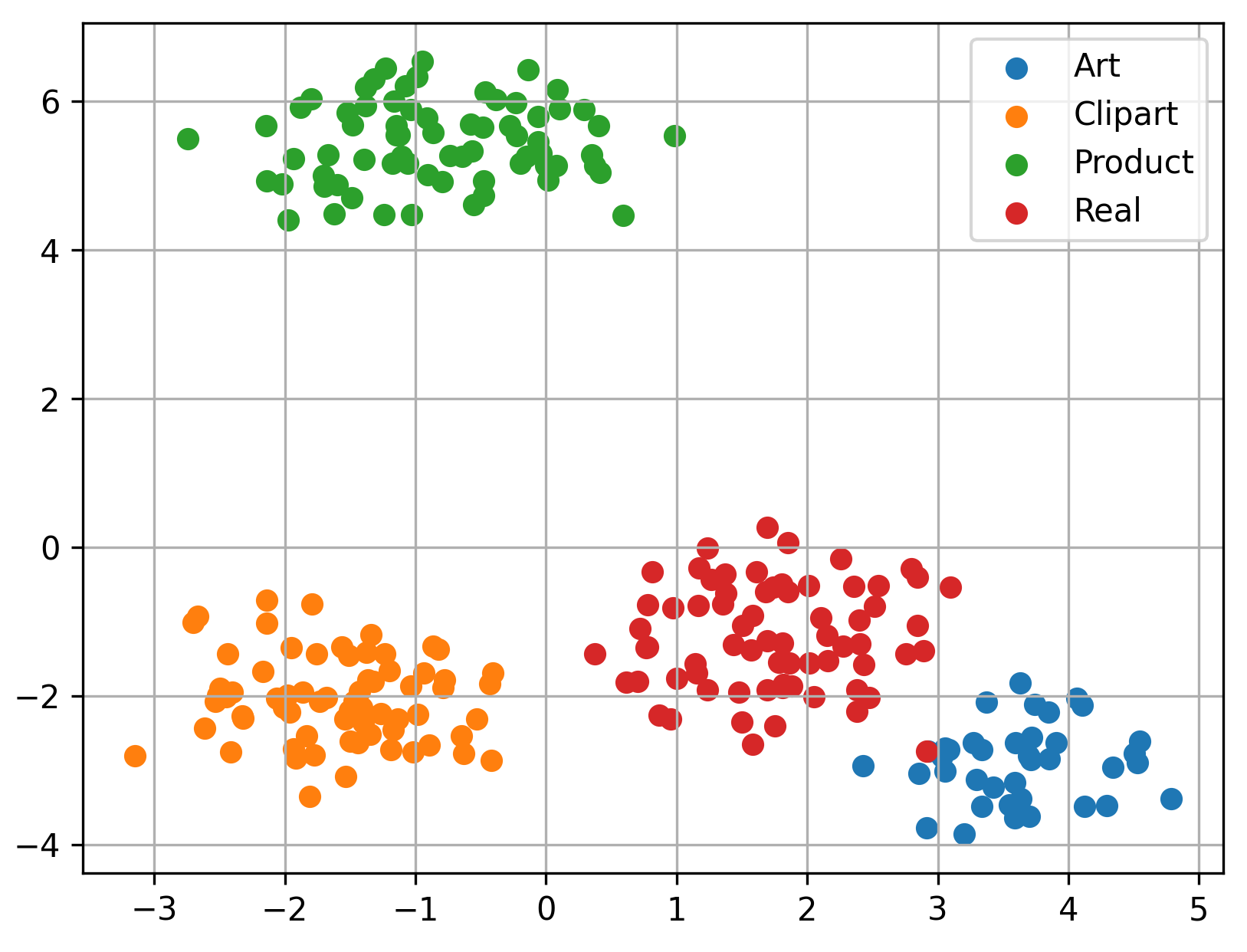} 
\caption{\textbf{BN statistics for domain separation.} t-SNE diagram of the batch-norm means extracted from the last CNN layer of ResNet-18 model. Clearly, the extracted batch-norm statistics allow for a clear separation between domains.}
\label{fig:tsne}
\vspace{-6pt}
\end{figure}

\subsection{Problem Setting}

Consider a model $f_{{\theta}_{0}}$ pre-trained on the source data $\mathcal{X}_{s} \sim \mathcal{D}_s$, where $\mathcal{D}_s$ denotes the source distribution. During deployment, the model encounters a sequence of test data $\mathcal{X}_{1} \rightarrow \mathcal{X}_{2}\rightarrow \ldots \rightarrow \mathcal{X}_{t} \rightarrow \ldots$, where $\mathcal{X}_{t}$ represents a batch of test samples from the test distribution $\mathcal{D}^t_{\textit{test}}$. Following the TTA setting, the model needs to adapt to each incoming test batch $\mathcal{X}_{t}$ and update its parameters from $f_{{\theta}_{t-1}} \rightarrow f_{{\theta}_{t}}$ in order to improve its predictions on the subsequent test batch $\mathcal{X}_{t+1}$. Since $\mathcal{D}^t_{\textit{test}}$ changes continually over time, our objective is to determine the specific time instance $t$ when a change in the domain occurs, $\mathcal{D}^t_{\textit{test}} \neq \mathcal{D}^{t-1}_{\textit{test}}$. Detecting these domain changes allows us to mitigate forgetting and error accumulation by reverting the model parameters to $f_{{\theta}_{0}}$, preserving source knowledge and enabling effective adaptation to new domains.


\subsection{Overall Framework}

\begin{figure*}[t!]
\centering
\includegraphics[width=1\textwidth]{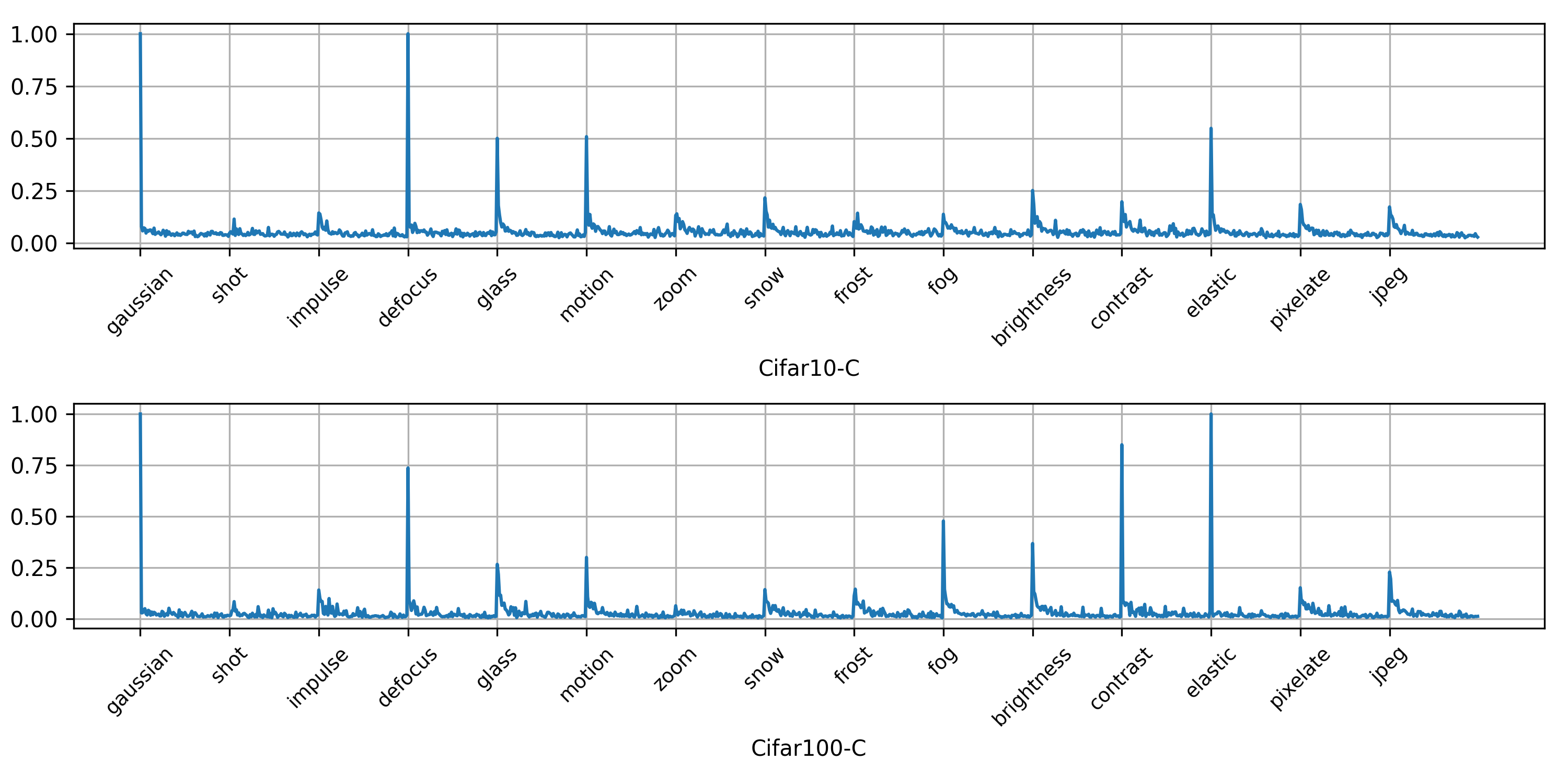} 
\caption{Plot of $\bar{\alpha}^{t}$ with respect to incoming target batches from 15 sequential target domains as the running statistics get aligned. The target datasets are from CIFAR10-C and CIFAR100-C. As the oracle-models (WideResNet-28 and ResNeXt-29) encounter test batches from a new domain, the running statistics gradually align with the global statistics of that domain. Consequently, during a domain change, $\bar{\alpha}^{t}$ exhibits a peak in its neighborhood due to the high KL divergence between the running statistics and the statistics of the incoming new-domain target batch. By detecting these peaks, it is possible to identify the occurrence of a domain change in the test batch. More plots of $\bar{\alpha}^{t}$ on other datasets can be found in the supplementary.}
\label{fig:plot_alpha}
\end{figure*}

In this work, we hypothesize that leveraging the batch-norm (BN) statistics of incoming test batches can provide valuable insights into domain differentiation. To illustrate this, we conduct an experiment using a pre-trained ResNet-18 model on ImageNet. By extracting the batch-norm mean from the last CNN layer for images from various domains in the Office-Home \cite{venkateswara2017deep} dataset, we create a t-SNE \cite{van2008visualizing} visualization (Figure \ref{fig:tsne}). The plot clearly shows distinct separations between domains, indicating that BN statistics exhibit unique patterns for different domains. Tracking changes in these statistics allows us to detect domain shifts. However, in dynamic environments with limited data availability, accurately estimating BN statistics becomes challenging.
 

Towards solving the problem, we aim to capture the global feature statistics of the current domain from the sequential incoming test batches, each containing limited data. When a domain shift occurs, the underlying distribution of the data changes significantly, causing a noticeable deviation in the feature statistics. By detecting these abrupt changes, we can identify the occurrence of a domain change. For this purpose, we employ a running average strategy to progressively correct and update the BN statistics to the incoming domain. Specifically, we first make a copy of the pre-trained source model that essentially serves as the oracle. During testing on the current batch $t$, let $\tilde{\mu}_{l_{c}}$ and $\tilde{\sigma}_{l_{c}}^{2}$ represent the BN statistics (mean and variance) extracted from the $l$-th layer's $c$-th channel of the oracle-model. We maintain two running averages as follows:
\begin{align}
    \mu_{l_{c}}^{t} &= (1-\bar{\alpha}^{t})\times\mu_{l_{c}}^{t-1} + \bar{\alpha}^{t}\times \tilde{\mu}_{l_{c}} \label{eq:running_mu}\\
    (\sigma_{l_{c}}^{t})^2 &= (1-\bar{\alpha}^{t})\times(\sigma_{l_{c}}^{t-1})^2 + \bar{\alpha}^{t}\times\tilde{\sigma}_{l_{c}}^{2} \ . \label{eq:running_sigma}
\end{align}
Here, $\mu_{l_{c}}^{t}$ and $(\sigma_{l_{c}}^{t})^2$ denote the running BN statistics that get updated with each incoming test batch. We initialize $\mu_{l_{c}}^{0}$ and $(\sigma_{l_{c}}^{0})^2$ with the BN statistics of the pre-trained source model. $\bar{\alpha}^{t}$ is the adaptive momentum value that controls the influence of the current batch on the running average. We want this value to gradually decrease over time when encountering test batches from the same domain. This decrease reflects the alignment between the running statistics and the global feature statistics of that specific domain. As the running statistics gradually align through incoming test data, the need for significant adjustments diminishes, and the running statistics better capture the underlying distribution of the domain. However, when a domain change occurs, the feature statistics of the current test batch deviate significantly from the running statistics. In such cases, the adaptive momentum $\bar{\alpha}^{t}$ should assume a higher weight, indicating the necessity for a larger correction to align the running feature statistics with the new domain.

To design this momentum term, we propose to use the KL divergence as a metric to quantify the domain shift. Assuming the batch statistics per channel of the BN layers as a univariate Gaussian distribution, we calculate the divergence between the running statistics up to the current time instance (approximated as ${\cal{N}}(\mu_{l_{c}}^{t-1},(\sigma_{l_{c}}^{t-1})^2)$) and the incoming test batch-norm statistics (approximated as ${\cal{N}}(\tilde{\mu}_{l_{c}},(\tilde{\sigma}_{l_{c}})^2)$). This allows us to obtain the \emph{unnormalized} value of the adaptive momentum for each layer as follows,
\begin{equation}
\begin{split}
    \alpha_{l}^{t} & = \frac{1}{C_{l}} \sum_{c=1}^{C_{l}}\mathcal{D}_{KL} \left [ \mathcal{N}\left(\mu_{l_{c}}^{t-1},(\sigma_{l_{c}}^{t-1})^2\right), \mathcal{N}\left(\tilde{\mu}_{l_{c}},\tilde{\sigma}_{l_{c}}^2\right)\right] 
                    \\ & = \frac{1}{C_{l}}\sum_{c=1}^{C_{l}} \log \left (\frac{\tilde{\sigma}_{l_{c}}}{\sigma_{l_{c}}^{t-1}} \right)
                   {+} \frac{\left(\sigma_{l_{c}}^{t-1}\right)^2{+}\left(\mu_{l_{c}}^{t-1}-\tilde{\mu}_{l_{c}}\right)^2}{2 \tilde{\sigma}_{l_{c}}^2}  {-} \frac{1}{2} \ 
 .\label{eq:alpha_per_layer}
\end{split}
\end{equation}
Where $C_{l}$ is the number of channels in $l$-th layer. Finally, the aggregated value across all layers, $\alpha^{t}$, is calculated as follows, 
\begin{align}
\alpha^{t} = \frac{\sum_{l} \alpha_{l}^{t}}{L} \  \label{eq:alpha_final}
\end{align}
where $L$ is the total number of layers. We track the highest value of the KL divergence to normalize each $\alpha^{t}$ to $\bar{\alpha}^{t}$, ensuring that the values remain within the range of 0 to 1. To initialize the highest domain distance value, we use the KL divergence between the pre-trained BN statistics and the statistics of the first target test batch. If a test batch yields an even higher KL divergence, this value is updated accordingly. Thus, $\bar{\alpha}^{t}$ assumes a value of 1 initially.

Furthermore, it is not necessary to consider all layers of the oracle-model to compute the momentum values. Recent work \cite{lim2023ttn} suggests that the penultimate layers are more sensitive to domain shifts. In our experiments, we only consider the layers of the last block in the model for the calculation. 


\begin{algorithm}
\caption{Overall Framework}
\label{alg:adaptive_momentum}
\begin{algorithmic}[1]
\REQUIRE Source model $f_{\theta_{0}}$, Oracle-model $f_{\theta_{orc}}$
\STATE Initialize $\mu_{l_{c}}^{0}$ and $(\sigma_{l_{c}}^{0})^2$ of the oracle-model with source pre-trained BN-statistics and set $\alpha_{max} \to 0^{+}$ 
\FOR{each incoming test batch $t$}
    \FOR{each layer $l$}
        \STATE Calculate adaptive momentum $\alpha_{l}^{t}$ for layer $l$ using Eqn. \ref{eq:alpha_per_layer}
    \ENDFOR
    \STATE Calculate mean adaptive momentum $\alpha^{t}$ of $L$ layers using Eqn. \ref{eq:alpha_final}
    \IF{ $\alpha^{t} > \alpha_{max}$ }
        \STATE $\alpha_{max} = \alpha^{t}$
    \ENDIF
    \STATE Normalize to get $\bar{\alpha}^{t} = \frac{\alpha^{t}}{\alpha_{max}}$ 
    \FOR{each layer $l$}
        \FOR{each channel $c$ in layer $l$}
            \STATE Update running averages using Eqn. \ref{eq:running_mu} and \ref{eq:running_sigma}
        \ENDFOR
    \ENDFOR
    
    \STATE Input the $\bar{\alpha}^{t}$ to the peak detection algorithm (Algorithm 2 in the Supplementary)
    \IF{peak}
        \STATE Restore source model parameters to $f_{\theta_{0}}$ 
        
    \ENDIF 
    \STATE Predict on test batch $t$ 
    \STATE Adapt model from $f_{{\theta}_{t-1}}$ to $f_{{\theta}_{t}}$ with selected TTA algorithm
\ENDFOR
\end{algorithmic}
\end{algorithm}

In summary, we propose an effective approach that gradually aligns the running statistics of each test batch with the global statistics of the underlying domain. This enables effective detection of domain shifts, as the running statistics undergo significant changes when a new domain is encountered. This can be trivially incorporated into any existing TTA algorithm to reset the model parameters whenever a new domain is encountered, as illustrated in Algorithm \ref{alg:adaptive_momentum}.

\subsection{Peak Detection}


Figure \ref{fig:plot_alpha} plots $\bar{\alpha}^{t}$ for incoming batches in our experiments. The value of $\bar{\alpha}^{t}$ exhibits a significant increase whenever a domain change occurs, reaching a peak in its neighborhood. As the model encounters more test batches from the same domain, the running statistics gradually align with the statistics of the domain, resulting in a decrease in the KL divergence. This leads to a decrease in the momentum value. By monitoring the online trend of $\bar{\alpha}^{t}$ and identifying the peak, we can detect domain changes in real time.


We utilize the online z-score algorithm \cite{kejriwal2015real} for peak detection. This algorithm maintains an exponential running average of the mean and standard deviation of inputs using a sliding window. For a query data point, it calculates the z-score, which represents the number of standard deviations the data point deviates from the running mean. An anomaly or peak is detected when the z-score exceeds a predefined threshold. The details of the algorithm is provided in the supplementary.



This is to be noted that the peak detection algorithm exclusively operates on the oracle model, while the original source model is consistently adapted (using any user preferred TTA algorithm) with the arrival of test batches. Whenever the oracle model identifies a domain shift, it restores the adapted source model to its original state, and the process of dynamic adaptation continues.

\section{Experiments}


\newcolumntype{R}[2]{%
    >{\adjustbox{angle=#1,lap=\width-(#2)}\bgroup}%
    l%
    <{\egroup}%
}

\newcommand*\rot{\multicolumn{1}{R{45}{1em}}}%

\begin{table*}[ht]
\centering
\footnotesize
\caption{Classification error rate $\downarrow$ (in \%) for the standard CIFAR100-to-CIFAR100C and ImageNet to ImageNet-C continual test-time adaptation task on corruption severity level $5$. Note that the “online” methods represent best-case scenarios that require manual restoration to source weight upon a domain change, and they serve as upper bounds for their corresponding continuous
adaptation counterparts. Online model rows are highlighted to emphasize best-case scenario.}
\label{tab:cifar100c}
\begin{tabular}{c|c|cccccccccccccccc}
\hline
                            & Time     & \multicolumn{3}{c}{t ----------------\textgreater{}}                                                 &             &            &             &           &           &            &          &                 &               &              &               &           &      \\ \cline{2-18} 
Dataset                     & Domains        & \rot{Gaussian} & \rot{Shot} & \rot{Impulse} & \rot{Defocus} & \rot{Glass} & \rot{Motion} & \rot{Zoom} & \rot{Snow} & \rot{Frost} & \rot{Fog} & \rot{Brightness} & \rot{Contrast} & \rot{Elastic} & \rot{Pixelate} & \rot{JPEG} & Mean \\ \hline
\multirow{6}{*}{\rotatebox{90}{CIFAR100C}}  & \mycc{DUA Online}  & \mycc{43.7}          & \mycc{39.8}      & \mycc{42.8}         & \mycc{33.1}        & \mycc{44.4}       & \mycc{30.5}        & \mycc{27.8}      & \mycc{34.9}      & \mycc{33.5}       & \mycc{43.6}     & \mycc{25.8}            & \mycc{32.3}          & \mycc{39.9}         & \mycc{41.4}          & \mycc{41.3}      & \mycc{37.0} \\
                            & DUA Continual           & 43.7          & \textbf{39.5}      & \textbf{42.2}         & 56.8        & 47.6       & 40.2        & 33.6      & 39.9      & 36.8       & 48.2     & 28.7            & 43.2          & 49.5         & 77.1          & 47.1      & 44.9 \\
                            & DUA + Ours    & \textbf{43.7}          & 39.7      & 42.7         & \textbf{33.1}        & \textbf{44.1}       & \textbf{30.4}        & \textbf{29.0}      & \textbf{34.5}      & \textbf{33.5}       & \textbf{43.8}     & \textbf{25.9}            & \textbf{32.5}          & \textbf{39.8}         & \textbf{41.4}          & \textbf{41.0}      & \textbf{37.0} \\
                            & \mycc{Tent Online} & \mycc{37.5}          & \mycc{35.3}      & \mycc{32.7}         & \mycc{25.7}        & \mycc{37.5}       & \mycc{27.5}        & \mycc{25.8}      & \mycc{30.6}      & \mycc{32.3}       & \mycc{33.1}     & \mycc{24.4}            & \mycc{28.1}          & \mycc{33.2}         & \mycc{28.7}          & \mycc{37.3}      & \mycc{31.3} \\
                            & Tent Continual          & \textbf{37.5}          & 37.1      & 44.3         & 41.3        & 56.5       & 55.6        & 57.9      & 69.7      & 75.0       & 83.3     & 86.2            & 93.7          & 95.5         & 96.2          & 96.9      & 68.4 \\
                            & 
                            Tent + Ours   & 37.8          & \textbf{35.5}      & \textbf{32.7}         & \textbf{25.8}        & \textbf{38.0}       & \textbf{27.6}        & \textbf{26.9}      & \textbf{30.8}      & \textbf{32.4}       & \textbf{33.9}     & \textbf{24.7}            & \textbf{28.3}          & \textbf{33.2}         & \textbf{28.8}          & \textbf{37.9}      & \textbf{31.6} \\ \hline
\multirow{6}{*}{\rotatebox{90}{ImageNet-C}} & \mycc{DUA Online}  & \mycc{85.7}          & \mycc{84.5}      & \mycc{84.6}         & \mycc{96.5}        & \mycc{88.3}       & \mycc{75.9}        & \mycc{61.9}      & \mycc{69.2}      & \mycc{68.7}       & \mycc{60.2}     & \mycc{35.6}            & \mycc{84.0}          & \mycc{60.6}         & \mycc{51.4}          & \mycc{60.3}      & \mycc{71.2} \\
                            & DUA Continual           & \textbf{85.7}          & 88.5      & \textbf{82.5}         & 99.8        & 99.8       & 98.8        & 89.9      & 88.6      & 79.8       & 84.4     & 49.5            & 96.3          & 78.5         & 71.5          & 60.3      & 83.6 \\
                            & DUA + Ours    & 85.7          & \textbf{84.6}      & 83.0         & \textbf{99.4}        & \textbf{90.8}       & \textbf{76.3}        & \textbf{62.1}      & \textbf{69.4}      & \textbf{68.9}       & \textbf{61.7}     & \textbf{35.7}            & \textbf{84.5}          & \textbf{62.8}         & \textbf{52.2}          & \textbf{59.9}      & \textbf{71.8} \\
                            & \mycc{Tent Online} & \mycc{74.7}          & \mycc{71.4}      & \mycc{73.8}         & \mycc{75.5}        & \mycc{75.4}       & \mycc{61.9}        & \mycc{52.6}      & \mycc{54.5}      & \mycc{61.3}       & \mycc{44.0}     & \mycc{33.8}            & \mycc{79.3}          & \mycc{46.5}         & \mycc{43.6}          & \mycc{50.1}      & \mycc{59.9} \\
                            & Tent Continual          & \textbf{74.7}          & \textbf{71.7}      & \textbf{69.7}         & 76.3        & 75.0       & 70.4        & 62.0      & 68.8      & 71.0       & 62.7     & 51.1            & 82.5          & 67.8         & 66.9          & 71.4      & 69.5 \\
                            & Tent + Ours   & 74.8          & 73.6      & 72.3         & \textbf{75.4}        & \textbf{74.9}       & \textbf{62.5}        & \textbf{52.7}      & \textbf{55.1}      & \textbf{62.3}       & \textbf{44.0}     & \textbf{33.6}            & \textbf{79.3}          & \textbf{47.7}         & \textbf{43.5}          & \textbf{50.0}      & \textbf{60.1} \\ \hline
\end{tabular}
\end{table*}

\subsection{Datasets}
\label{sec:datasets}

 \noindent $\bullet$ \textbf{CIFAR10C, CIFAR100C, and ImageNet-C}: CIFAR10 and CIFAR100 \cite{krizhevsky2009learning} are popular image classification datasets consisting of 10,000 test images. To evaluate the robustness of trained models, CIFAR10C and CIFAR100C \cite{hendrycks2019benchmarking} were developed. These datasets introduce 15 distinct types of noise at varying severity levels (1 to 5) to the original CIFAR10 and CIFAR100 test images. Similarly, ImageNet-C \cite{hendrycks2019benchmarking} is the noisy counterpart of the ImageNet dataset \cite{deng2009imagenet}. These datasets serve as widely-used benchmarks in the field of continual TTA \cite{wang2022continual}.
\newline
 $\bullet$ \textbf{Digits:} The Digit benchmark is a standard dataset for digit classification, comprising ten classes. For our experiments, we utilized five domains: MNIST (MT), USPS (UP), SVHN (SV), MNIST-M (MM), and Synthetic Digits (SY).
 \newline
 $\bullet$ \textbf{Office-Home:}: The Office-Home dataset \cite{venkateswara2017deep} comprises four domains, namely Art (Ar), Clipart (Cl), Product (Pr), and Real World (Re), each containing 65 classes. 
 \newline
$\bullet$ \textbf{Cityscapes to ACDC:} Cityscapes \cite{cordts2016cityscapes} is a large-scale dataset that has dense pixel-level annotations for 30 classes grouped into 8 categories. The Adverse Conditions Dataset \cite{sakaridis2021acdc} has images corresponding to fog, night-time, rain, and snow weather conditions. The number of classes is the same as the evaluation classes of the Cityscapes dataset. Keeping accordance with the standard setting, we evaluate our model on $19$ semantic labels without considering the void label.


\subsection{Baseline Methods}
We utilize TENT \cite{wang2020tent} as the primary adaptation method and incorporate it with our proposed approach on detecting domain changes and restoring model parameters. We selected TENT primarily due to its lightweight and efficient nature, as it does not require additional modules and can perform adaptation with a single back-propagation step. We also compare our method with a variant of TENT called `TENT-Online' \cite{wang2022continual}. TENT-Online assumes the availability of an oracle that can detect domain changes and resets the model accordingly. This serves as an upper-bound comparison for our method, representing the best-case scenario. However, in practice, such an oracle is not practical as this requires additional ground truth domain knowledge. We show that our method achieves comparable results to TENT-Online without the need for such ground truth knowledge.


Apart from TENT, we also integrate our method with DUA \cite{mirza2022norm} to illustrate the versatility of the proposed domain change detection module with respect to the underlying TTA algorithm. By detecting domain changes and restoring DUA parameters, we extend its capability to perform adaptation in dynamic environments. We also compare our results with CoTTA \cite{wang2022continual} and EATA \cite{niu2022efficient}. These two methods specifically focus on continual test time adaptation. We use the official implementations for all baselines.

\begin{table*}[t!]
\footnotesize
\caption{Classification error rate $\downarrow$ (in \%) for the standard CIFAR100-to-CIFAR100C and ImageNet to ImageNet-C adaptation task on corruption severity level $5$ against current state-of-the-art models on continual test time adaptation.}
\label{tab:sota}
\begin{tabular}{c|c|cccccccccccccccc}
\hline
\multirow{2}{*}{Dataset}    & Time        & \multicolumn{3}{c}{t ----------------\textgreater{}}                                           &                                &                              &                               &                             &                             &                              &                            &                               &                                 &                                &                                 &                                                  &      \\ \cline{2-18} 
                            & Domains     & \rot{Gaussian} & \rot{Shot} & \rot{Impulse} & \rot{Defocus} & \rot{Glass} & \rot{Motion} & \rot{Zoom} & \rot{Snow} & \rot{Frost} & \rot{Fog} & \rot{Bright} & \rot{Contrast} & \rot{Elastic} & \rot{Pixelate} & \rot{JPEG} & {Mean}
                             \\ \cline{1-18}
\multirow{5}{*}{\rotatebox{90}{CIFAR100C}} & Source      & 72.3                            & 67.4                        & 39.0                           & 29.4                           & 53.6                         & 30.5                          & 28.8                        & 39.1                        & 45.5                         & 50.3                       & 29.7                          & 55.4                            & 37.2                           & 74.8                            & \multicolumn{1}{c}{41.0}                        & {46.3} \\
                            & BN-Stat \cite{schneider2020improving}     & 42.1                            & 40.7                        & 42.7                           & 27.6                           & 41.9                         & 29.7                          & 27.9                        & 34.9                        & 35                           & 41.5                       & 26.5                          & 30.3                            & 35.7                           & 32.9                            & \multicolumn{1}{c}{41.2}                        & {35.4} \\
                            & EATA \cite{niu2022efficient}       & 39.7                            & 37.2                        & 37.0                           & 26.9                           & 40.0                         & 28.4                          & 26.5                        & 32.0                        & 32.9                         & 38.2                       & 25.2                          & 29.9                            & 34.3                           & 30.5                            & \multicolumn{1}{c}{39.0}                        & {33.2} \\
                            & CoTTA \cite{wang2022continual}      & 40.1                            & 37.7                        & 39.7                           & 26.9                           & 38.0                         & 27.9                          & \textbf{26.4}                        & 32.8                        & \textbf{31.8}                         & 40.3                       & 24.7                          & \textbf{26.9}                            & \textbf{32.5}                           & \textbf{28.3}                            & \multicolumn{1}{c}{\textbf{33.5}}                        & {32.5} \\
                                & Tent + Ours & \textbf{37.8}                            & \textbf{35.5}                        & \textbf{32.7}                           & \textbf{25.8}                           & \textbf{38.0}                         & \textbf{27.6}                          & 26.9                        & \textbf{30.8}                        & 32.4                         & \textbf{33.9}                       & \textbf{24.7}                          & 28.3                            & 33.2                           & 28.8                            & \multicolumn{1}{c}{37.9}                        & {\textbf{31.6}} \\ \hline
\multirow{5}{*}{\rotatebox{90}{ImageNetC}} & Source      & 97.9                            & 96.9                        & 98.2                           & 81.9                           & 89.7                         & 84.9                          & 78.2                        & 83.3                        & 77.3                         & 76.2                       & 41.2                          & 94.4                            & 82.9                           & 79.2                            & \multicolumn{1}{c}{68.7}                        & {82.1} \\
                            & BN-Stat \cite{schneider2020improving}     & 85                              & 83.7                        & 85                             & 84.7                           & 84.3                         & 73.7                          & 61.2                        & 66                          & 68.2                         & 52.1                       & 34.9                          & 82.7                            & 55.9                           & 51.3                            & \multicolumn{1}{c}{59.8}                        & {68.6} \\
                            & EATA \cite{niu2022efficient}      & 82.4                            & 76.9                        & 73.9                           & 77.4                           & \textbf{73.1}                         & 63.9                          & 54.0                        & 60.9                        & 61.2                         & 49.1                       & 36.0                          & 67.3                            & 49.4                           & 45.6                            & \multicolumn{1}{c}{49.9}                        & {61.4} \\
                            & CoTTA \cite{wang2022continual}      & 84.5                            & 81.9                        & 79.8                           & 80.8                           & 78.2                         & 67.3                          & 57.6                        & 60.5                        & \textbf{60.4}                         & 48.2                       & 36.5                          & \textbf{64.0}                            & \textbf{47.3}                           & \textbf{41.1}                            & \multicolumn{1}{c}{\textbf{45.2}}                        & {62.2} \\
                            & Tent + Ours & \textbf{74.8}                            & \textbf{73.6}                        & \textbf{72.3}                           & \textbf{75.4}                           & 74.9                         & \textbf{62.5}                          & \textbf{52.7}                        & \textbf{55.1}                        & 62.3                         & \textbf{44.0}                       & \textbf{33.6}                          & 79.3                            & 47.7                           & 43.5                            & \multicolumn{1}{c}{50.0}                        & {\textbf{60.1}} \\ \hline
\end{tabular}
\end{table*}

\subsection{Implementation Details}
For CIFAR100C and ImageNet-C experiments, we adopt the pre-trained ResNeXt-29 \cite{xie2017aggregated} and ResNet-50 respectively from Robustbench \cite{croce2020robustbench} for all methods. We use ResNet-18 for the Digit and Office-home experiments. Our batch size is set to $64$ for ImageNet-C and $128$ for other classification experiments. For the TENT approach adopted for our method, we use Adam optimizer with a learning rate of $0.001$. Similar to \cite{wang2022continual}, we utilize the validation set compiled by RobustBench for ImageNet-C dataset. For the peak detection algorithm, we use a sliding window of $10$ and a momentum of $0.1$ for incoming values. The threshold is taken as $15$ standard deviations. We maintain consistency with the respective papers of the compared methods by using the same learning rate, optimizer, and rest of the hyperparameters. For all the comparing methods we use their official implementations. 


\subsection{Experiments on CIFAR100C and ImageNet-C}

We first evaluate our method on CIFAR100C and ImageNet-C. Particularly, given a pre-trained model on CIFAR100/ImageNet, we adapt the model sequentially to 15 types of unseen domains/noise sets. Each of the noise sets has a total of $10,000$ images. We show the results in Table \ref{tab:cifar100c}. DUA-Online and TENT-Online models are manually reset when there is a domain change and thus, the models act as an upper-bound for comparison with our method. On the other hand, TENT and DUA Continual are continually adapted (lifelong) to test data without manual resetting. It can be observed from the table that the performance of both TENT and DUA deteriorates over time, as the models continually adapt to unseen test samples. In the case of ImageNet-C the deterioration is much more prominent because ImageNet-C has a total of $1000$ classes, which makes it difficult for the model to produce reliable pseudo-labels and thus contributing to error accumulation. The deviation of the results from the corresponding online models also highlights the poor performance of TENT and DUA. The performance in fact deteriorates over time in comparison to their corresponding online models due to catastrophic forgetting and gradual error accumulation. On the other hand, when our method is added to TENT and DUA in order to automatically detect a domain change and reset the model parameters accordingly, the performance of both models improves by a big margin. The results also get very much close to their corresponding online models which verifies that our method can effectively mimic the online models without requiring any domain knowledge. 

\subsection{Experiments on Digits and Office-Home}
We next perform experiments on digit and officehome datasets. In order to simulate a dynamic environment, we train a model on the train set of one dataset, and sequentially adapt on the test sets of rest of the datasets for a total of $10$ cycles. The results are shown in Table \ref{tab:digit} and Table \ref{tab:office-home}. The columns in the tables show the dataset in which the source model is trained on. For example, in case of the `MM' column of \ref{tab:digit}, the source model is trained on MNIST-M dataset and then during test-time, the model is adapted to sequential unseen domains in the following order: MT $\rightarrow$ UP $\rightarrow$ SV  $\rightarrow$ SY. This whole cycle goes on for a total of $10$ times. Similarly, in case of the `Art' column of Table \ref{tab:office-home}, the source model is trained on Art dataset and then, the model is adapted sequentially to unseen domains in the following order: Cl $\rightarrow$ Pr $\rightarrow$ Rw for a total of $10$ times. The mean classification error over the whole sequence is reported at the tables. It can be observed from both the table that our method performs better than TENT and DUA Continual. Our results are very close to the online models, while in some cases, i.e., SYNDIG, SVHN, Pr it is equal. 

\begin{table}[t!]
\centering
\caption {Classification error rate on continual adaptation task on digit datasets. The column header represents the source dataset where the model is trained on. Mean classification error over all the cycles is reported here.}
\label{tab:digit}
\footnotesize
\begin{tabular}{ccccccc}
\hline
Method        & MM   & MT   & UP   & SV   & SY   & Avg \\ \hline
\mycc{DUA Online}  & \mycc{36.3} & \mycc{71.6} & \mycc{74.7} & \mycc{29.7} & \mycc{24.1} &   \mycc{47.3}  \\
DUA Continual          & 36.6 & 73.8 & 76.0 & 29.8 & 24.4 &   48.1  \\
DUA + Ours    & \textbf{36.4} & \textbf{71.8} & \textbf{74.7} & \textbf{29.7} & \textbf{24.1} &  \textbf{47.3}   \\ \hline
\mycc{Tent Online} & \mycc{37.1} & \mycc{72.5} & \mycc{75.4} & \mycc{27.7} & \mycc{22.5} &  \mycc{47.0}    \\
Tent Continual       & 43.1  & 87.9  & 86.0  & 28.1  & 23.4  & 53.7     \\
Tent + Ours   & \textbf{38.0} & \textbf{72.8} & \textbf{75.6} & \textbf{27.8} & \textbf{22.5} &  \textbf{47.3}  \\ \hline
\end{tabular}
\end{table}

\begin{table}[t!]
\centering
\caption {Classification error rate on continual adaptation task on office-home dataset.}
\label{tab:office-home}
\footnotesize
\begin{tabular}{cccccc}
\hline
Method        & Ar   & Cl   & Pr   & Rw   & Avg \\ \hline
\mycc{DUA Online}  & \mycc{46.7} & \mycc{47.0} & \mycc{51.5} & \mycc{41.7} &    \mycc{46.7} \\
DUA Continual          & 49.6 & 47.5 & 52.4 & 43.2 &  48.2   \\
DUA + Ours    & \textbf{47.0} & \textbf{47.2} & \textbf{52.1} & \textbf{41.9} &  \textbf{47.1}   \\ \hline
\mycc{Tent Online} & \mycc{48.1} & \mycc{46.4} & \mycc{51.4} & \mycc{41.8} &    \mycc{46.9} \\
Tent Continual        & 79.2 & 63.6 & 75.2 & 47.7 &  66.4   \\
Tent + Ours   & \textbf{48.9} & \textbf{47.3} & \textbf{51.4} & \textbf{42.1} &  \textbf{47.4}  \\ \hline
\end{tabular}
\end{table}

\subsection{Experiments on Cityscapes to ACDC}
\begin{table}[h!]
\footnotesize
\centering
\caption{Semantic segmentation results (mIoU in \%) on the Cityscapes-to-ACDC online continual test-time adaptation task. We evaluate the four test conditions continually for ten times and report the mean here. `O' refers to the online model.}
\label{tab:segmentation}
\begin{tabular}{c|ccc|ccc}
\hline
Method & \mycc{DUA-O} & DUA-C  & +Ours & \mycc{Tent-O} & Tent-C & +Ours \\ \hline
mIoU   & \mycc{20.8}         & 20.5 & \textbf{20.7}     & \mycc{17.1}          & 14.7 & \textbf{17.0}      \\ \hline
\end{tabular}
\end{table}
In this experiment, we evaluate our method on the more complex continual test-time adaptation scenario - semantic segmentation of Cityscapes to ACDC. In order to simulate real-life situations where comparable environments may be encountered, we iterate an identical four condition sequence for a total of ten times, resulting in a total of 40 repetitions: Fog $\rightarrow$ Rain $\rightarrow$ Snow $\rightarrow$ Night $\rightarrow$ Fog $\dots$ Specifically, we train the model on the clean weather condition of Cityscapes dataset and adapt to this unseen weather condition sequence from ACDC. We use DeepLab v3+ \cite{chen2018encoder} with a ResNet-18 encoder for the experiment. The batch size used is $4$. We report the mean performance over all the sequences.


The results are shown in Table \ref{tab:segmentation}. It can be observed that the continual models perform worse than the online models. The performance gap between the two models is even more highlighted in the case of TENT. On the other hand, in both cases, adding our method helps DUA and TENT to reach performance on par with the online model. 

\subsection{Comparison with State-of-the-Art}
In this section, we evaluate our method against state-of-the-art approaches specifically designed for the continual adaptation task. We conduct experiments on CIFAR100-C and ImageNet-C. We compare our method with CoTTA \cite{wang2022continual} and EATA \cite{niu2022efficient}. CoTTA mitigates catastrophic forgetting by employing stochastic restoration of source weights, while EATA incorporates Fisher regularizer to limit drastic changes in important model parameters, thereby preserving source knowledge. In comparison, our method offers a more structured and intuitive approach to retaining source knowledge. Additionally, we also compare our method with BN-Stat \cite{schneider2020improving}, which replaces the pre-trained batch-norm statistics with statistics estimated from the test batch.

In the initial experiment, we conduct adaptation to the 15 corruptions from CIFAR100-C and ImageNet-C, each with a severity level of 5 \cite{wang2022continual}. The results are presented in Table \ref{tab:sota}. The table reveals that our method surpasses all the baseline models on both datasets in mean performance. Despite CoTTA and EATA utilizing advanced techniques such as mean teacher and advanced regularizers to retain source knowledge and mitigate error propagation, our simpler approach achieves superior performance compared to both these methods.

We next look at a more challenging adaptation scenario. For this experiment, we gradually change the severity level in CIFAR100C for each corruption set as follows: $1\rightarrow2\rightarrow3\rightarrow4\rightarrow5\rightarrow4\rightarrow3\rightarrow 2 \rightarrow 1$. Hence, we have a total of $15 \times 5$ or $75$ unseen domains. This specific experiment \cite{wang2022continual} is designed to highlight the catastrophic forgetting and error accumulation issue more prominently. 

\begin{table}[h!]
\footnotesize
\centering
\caption{Results for gradually changing noise level. The mean of the whole sequence is reported here.}
\begin{tabular}{c|ccc}
\hline
Method & EATA & CoTTA & Tent+Ours \\ \hline
Error Rate (in $\%$)  & 34.9 & 28.1  & \textbf{26.8}      \\ \hline
\end{tabular}
\end{table}

As can be seen from the Table, our method again outperforms both CoTTA and EATA in this challenging adaptation task. Notably, our method achieves an impressive $8.1\%$ improvement compared to EATA, further highlighting its effectiveness and superiority in mitigating catastrophic forgetting and error accumulation.

\subsection{Analysis on Forgetting}
\begin{figure}[t!]
\centering
\includegraphics[width=0.45\textwidth]{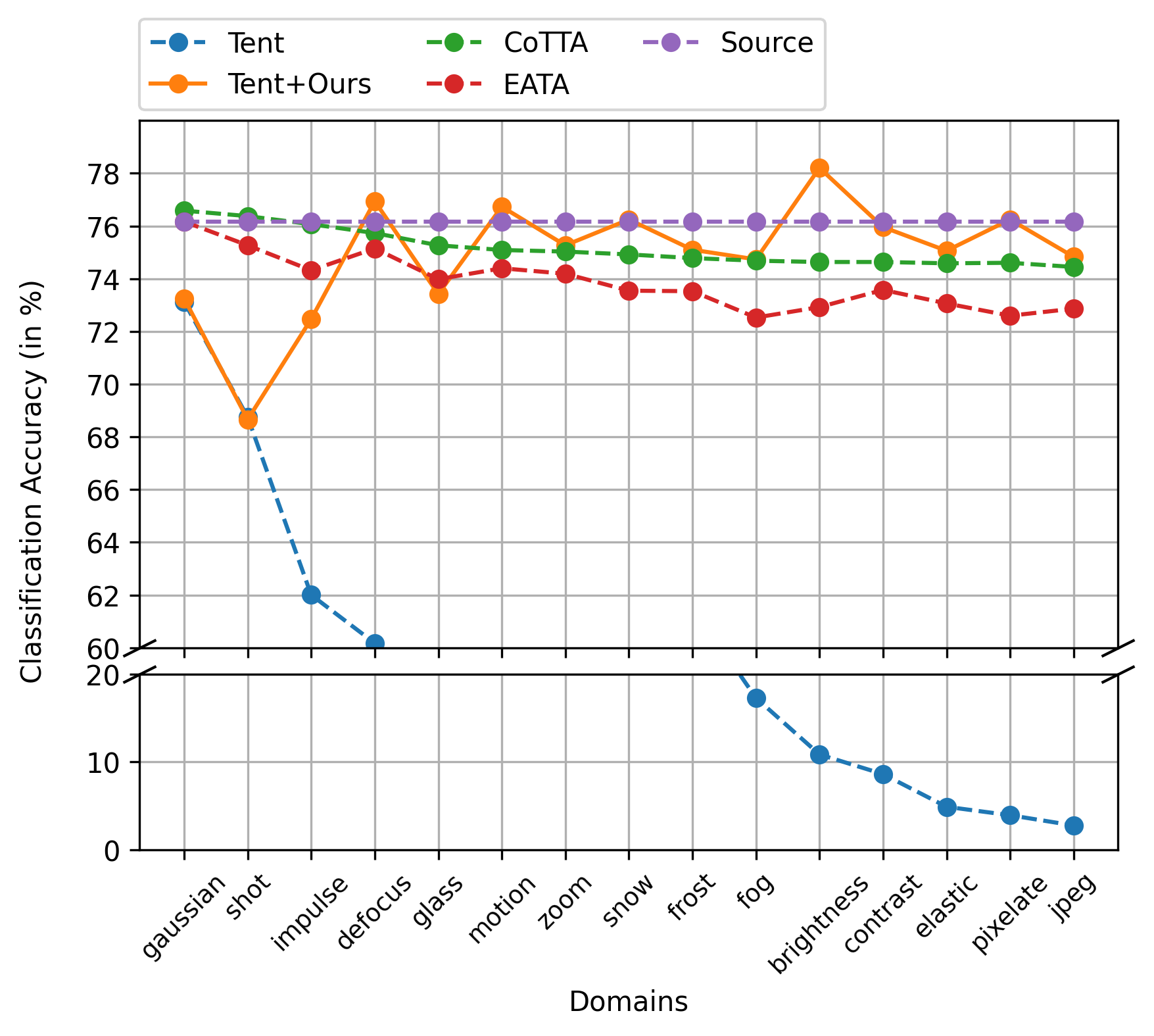} 
\caption{We assess the source knowledge by evaluating the methods on the source test set following the completion of adaptation to each domain of CIFAR-100C. Observing the results, it becomes evident that our method aligns more with the source accuracy, signifying robustness to the forgetting of source knowledge compared to SOTA methods.}
\label{fig:forgetting}
\end{figure}
In this section, we demonstrate the robustness of our method against catastrophic forgetting by evaluating the classification accuracy on the source test set after completing adaptation to each domain \cite{niu2022efficient}. Specifically, we take the CIFAR100-C dataset and after adapting the model to each of the $15$ unseen domains, we check the accuracy on the test set of the clean CIFAR100 dataset. This helps to quantify the reduction of source knowledge after each adaptation cycle. The results are shown in Figure \ref{fig:forgetting}.

From the figure, it is apparent that the test accuracy of the TENT method gradually deteriorates as it encounters new domains. This degradation of source knowledge directly correlates to poor generalization on target domains, as shown in Table \ref{tab:cifar100c}. For reference, the source accuracy is also plotted, representing the ideal scenario with the curve being completely flat. It can be observed that our method almost overlaps with the source accuracy curve. This shows that we achieve almost no forgetting of source knowledge with our method. 
CoTTA and EATA outperform TENT by a big margin because these methods are specifically designed to combat forgetting. Nevertheless, these two specialized methods also are not completely robust to forgetting, as observed by the gap from the flat source accuracy curve in the figure. Our method surpasses even CoTTA and EATA in terms of accuracy. Furthermore, we have carried out additional sensitivity analyses on the threshold value used in the peak detection algorithm to ensure its robustness. Detailed results, along with decision diagram of our method and an analysis of scenarios where the domain gaps between various domains are very small, can be found in the supplementary material.


\section{Conclusion}
This paper addresses the novel problem of detecting domain changes in dynamic environments. We estimate global batch-norm statistics of a domain using an adaptive momentum, which undergoes a significant change during distribution shifts. By detecting these domain changes and restoring model parameters to their source pre-trained values, we have shown that our method effectively mitigates the issues of catastrophic forgetting and error accumulation observed in traditional TTA methods. This enhances the robustness of TTA methods in dynamic environments, and ensures their performance is maintained over time.

\section*{Acknowledgement}
The work was partially supported by ONR grant N00014-19-1-2264 and the NSF grants CCF-2008020 and IIS-1724341.

{\small
\bibliographystyle{ieee_fullname}
\bibliography{egbib}
}

\end{document}


\title{Supplementary}  

\maketitle
\thispagestyle{empty}
\appendix


\section{Peak Detection Algorithm}

Here, we provide details of the Peak Detection Algorithm.  

\renewcommand{\thealgorithm}{2}
\begin{algorithm}
\caption{Online Peak Detection}
\label{alg:online_peak_detection}
\begin{algorithmic}
\REQUIRE $\text{window\_size}$: Size of sliding window
\REQUIRE $\text{threshold}$: Z-score threshold for peak detection
\REQUIRE $k$: Influence value for the running statistics
\REQUIRE $\text{data}$: Incoming $\bar{\alpha}^{t}$ stream
\FOR{each incoming $\bar{\alpha}^{t}$}
    \IF{sliding window length $\leq$ window\_size}
        \STATE Append $\bar{\alpha}^{t}$ to window  
        \IF {sliding window length $=$ window\_size}
            \STATE Initialize running $\mu_{r}$ and $\sigma_{r}$ as simple mean and standard deviation within sliding window
        \ENDIF
    \ELSE
        \STATE Calculate z-score for $\bar{\alpha}^{t}$: $z = \frac{\bar{\alpha}^{t} - \mu_{r}}{\sigma_{r}}$
        \IF{$z > \text{threshold}$}
            \STATE Consider $\bar{\alpha}^{t}$ as a peak or anomaly
        \ENDIF
        \STATE Slide the window by removing the oldest data point and appending $\bar{\alpha}^{t}$
        \STATE Calculate current mean $\mu_{c}$ and std $\sigma_{c}$ of the sliding window
        \STATE Update the running $\mu_r$ and $\sigma_r$ using the current $\mu_{c}$ and $\sigma_{c}$ with following eqns: \\
        $\mu_{r} = (1-k) \times \mu_{r} + k \times \mu_{c}$ \\
        $\sigma_{r} = (1-k) \times \sigma_{r} + k \times \sigma_{c}$
        
    \ENDIF
\ENDFOR
\end{algorithmic}
\end{algorithm}

\section{Additional $\bar{\alpha}^{t}$ Plots}
\begin{figure}[h!]
\centering
\includegraphics[width=0.53\textwidth]{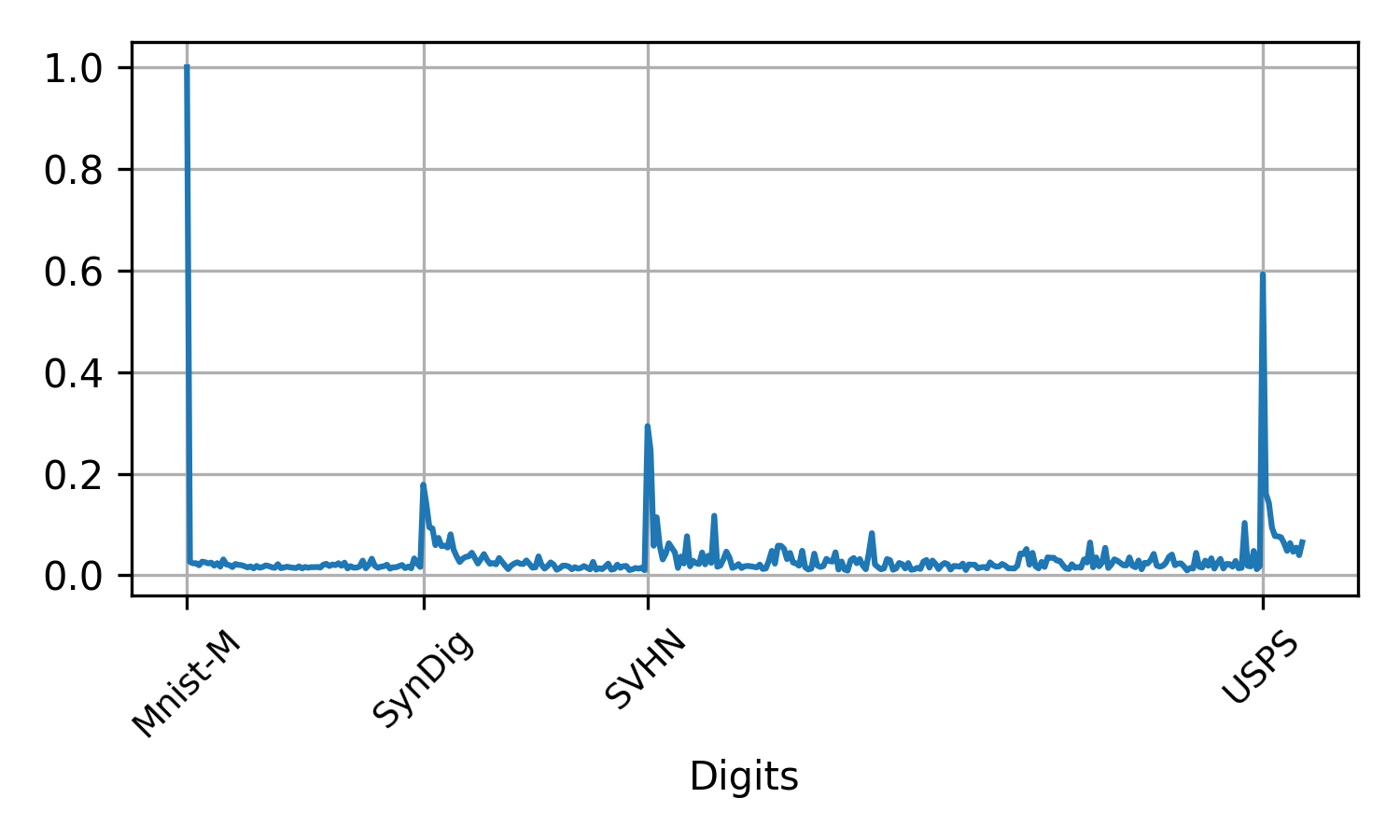} 
\caption{$\bar{\alpha}^{t}$ plot with respect to incoming test batches. The source model is trained on MNIST dataset and during test-time sequentially adapted to the remaining digit datasets.}
\label{fig:alpha_digit}
\end{figure}

\begin{figure}[h!]
\centering
\includegraphics[width=0.53\textwidth]{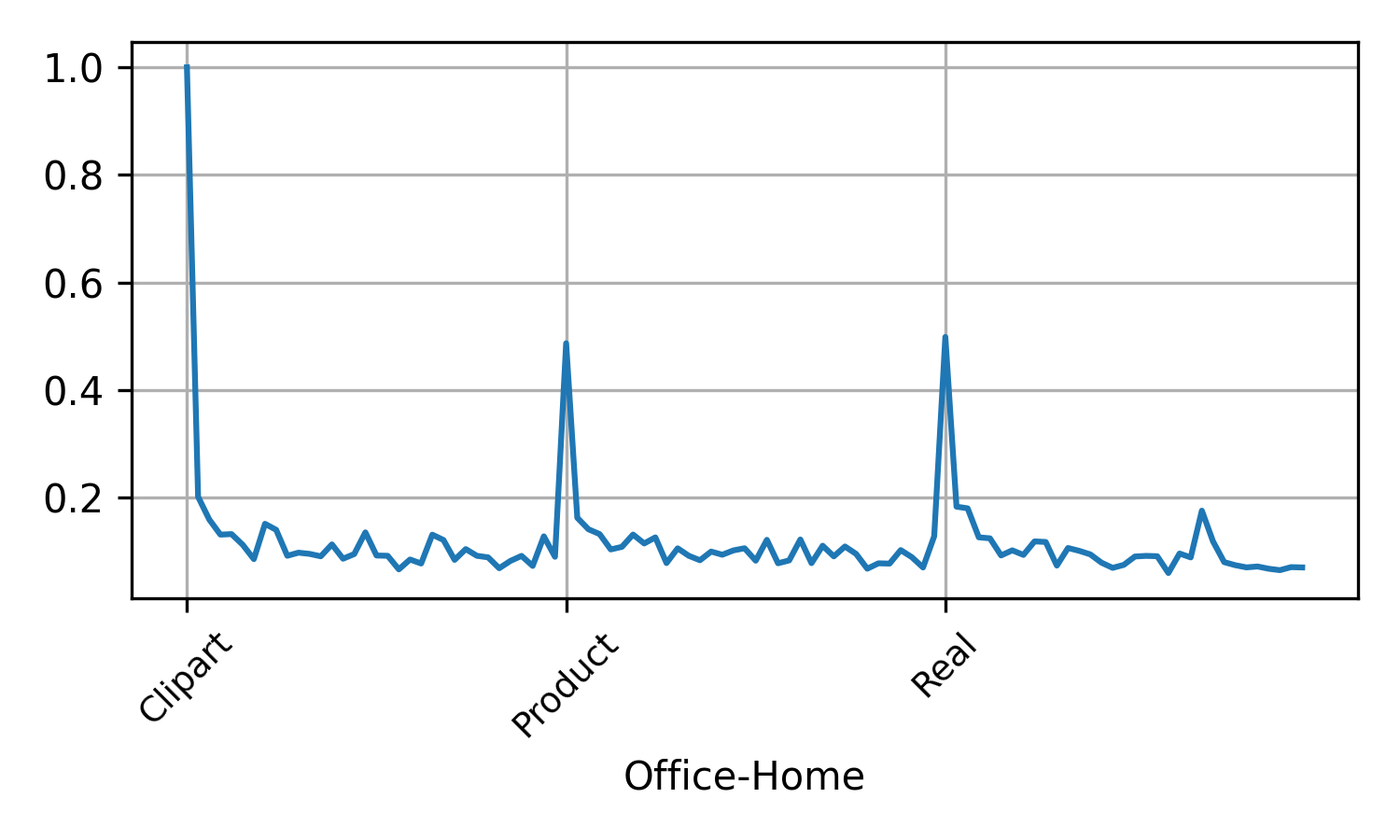} 
\caption{$\bar{\alpha}^{t}$ plot with respect to incoming test batches. The source model is trained on Art dataset and during test-time sequentially adapted to the remaining office-home datasets.}
\label{fig:alpha_office}
\end{figure}

\begin{figure*}[t!]
\centering
\includegraphics[width=1\textwidth]{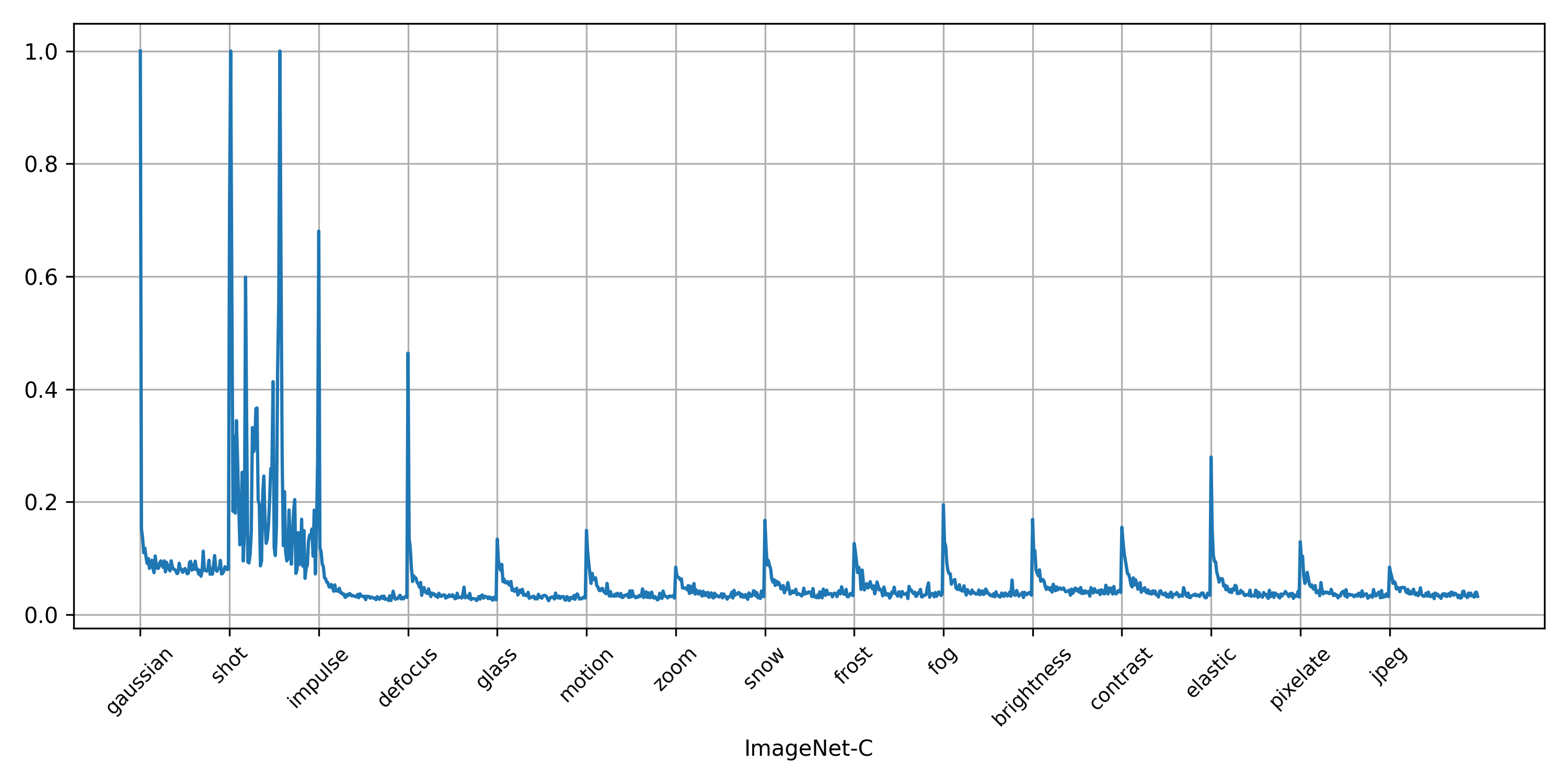} 
\vspace{-20pt}
\caption{$\bar{\alpha}^{t}$ plot with respect to incoming test batches for ImageNet-C dataset}
\label{fig:alpha_imagenet}
\end{figure*}

In this section, we present additional visualizations of the $\bar{\alpha}^{t}$ values obtained from our experiments. We first include visualizations of the $\bar{\alpha}^{t}$ with respect to incoming test batches for the first cycle of domains from the experiments of digit in Fig. \ref{fig:alpha_digit} and office-home in Fig. \ref{fig:alpha_office}. We also show the plot for ImageNet-C with severity level 5 in Fig. \ref{fig:alpha_imagenet}.

\section{Decision Diagram}
In this section, we visualize the domain change locations detected by our algorithm.

\begin{figure*}[t!]
\centering
\includegraphics[width=1\textwidth]{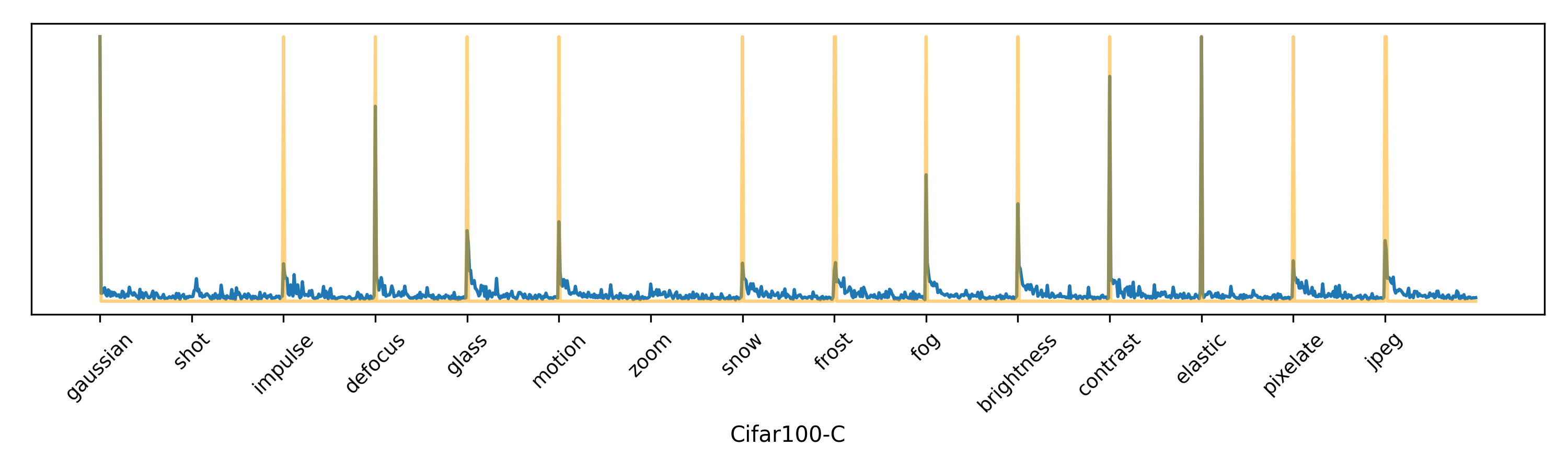} 
\vspace{-20pt}
\caption{Our algorithm's decision points are overlaid onto $\bar{\alpha}^{t}$ plot with respect to incoming test batches for CIFAR100-C dataset. Instances where our proposed algorithm detects a domain change are marked by orange lines.}
\label{fig:decision}
\end{figure*}

Fig. \ref{fig:decision} shows the visualization for CIFAR-100C. It can be observed that our method is capable of correctly detecting domain changes most of the time. It should be noted that, as our method rapidly aligns the running BN stats with the new domain, the false positive rates are also extremely low.

\section{Sensitivity Analysis}
The critical hyperparameter of the peak detection algorithm is the threshold value. We have used a value of 15 standard deviations for all our experiment. Here, we change the threshold value and plot the corresponding decision diagrams for CIFAR-100C.

\begin{table*}[t!]
\footnotesize
\caption{Classification error rate on CIFAR-100C with a severity level of 1. When the domain gap is low, the continual method's performance is on par with the online model. Nonetheless, our method managed to improve the continual method's performance even more.}
\label{tab:limitation}
\begin{tabular}{cccccccccccccccc|c}
\hline
\multicolumn{1}{c}{Method} & GN   & SN   & IN   & DB   & GB   & MB   & ZB   & Snow & Frost & Fog  & Bright & Contrast & Elastic & Pixel & JPEG & Mean \\ \hline
Tent Online                  & 26.2 & 24.1 & 22.8 & 22.2 & 30.7 & 23.5 & 22.9 & 23.6 & 24.1  & 22.1 & 21.9   & 22.6     & 26.7    & 23.7  & 28.7 & 24.4 \\
Tent Continual               & 26.2 & 24.9 & 24.6 & 24.1 & 31.8 & 26.5 & 26.2 & 26.7 & 26.7  & 25.9 & 25.8   & 25.9     & 29.2    & 26.2  & 30.1 & 26.7 \\
Tent+Ours                    & 26.4 & 25.2 & 22.9 & 22.5 & 30.6 & 23.5 & 23.2 & 23.7 & 24.9  & 24.0 & 23.9   & 23.8     & 28.5    & 25.2  & 30.4 & 25.2 \\ \hline
\end{tabular}
\end{table*}

\begin{figure*}[t!]
\centering
\includegraphics[width=1\textwidth]{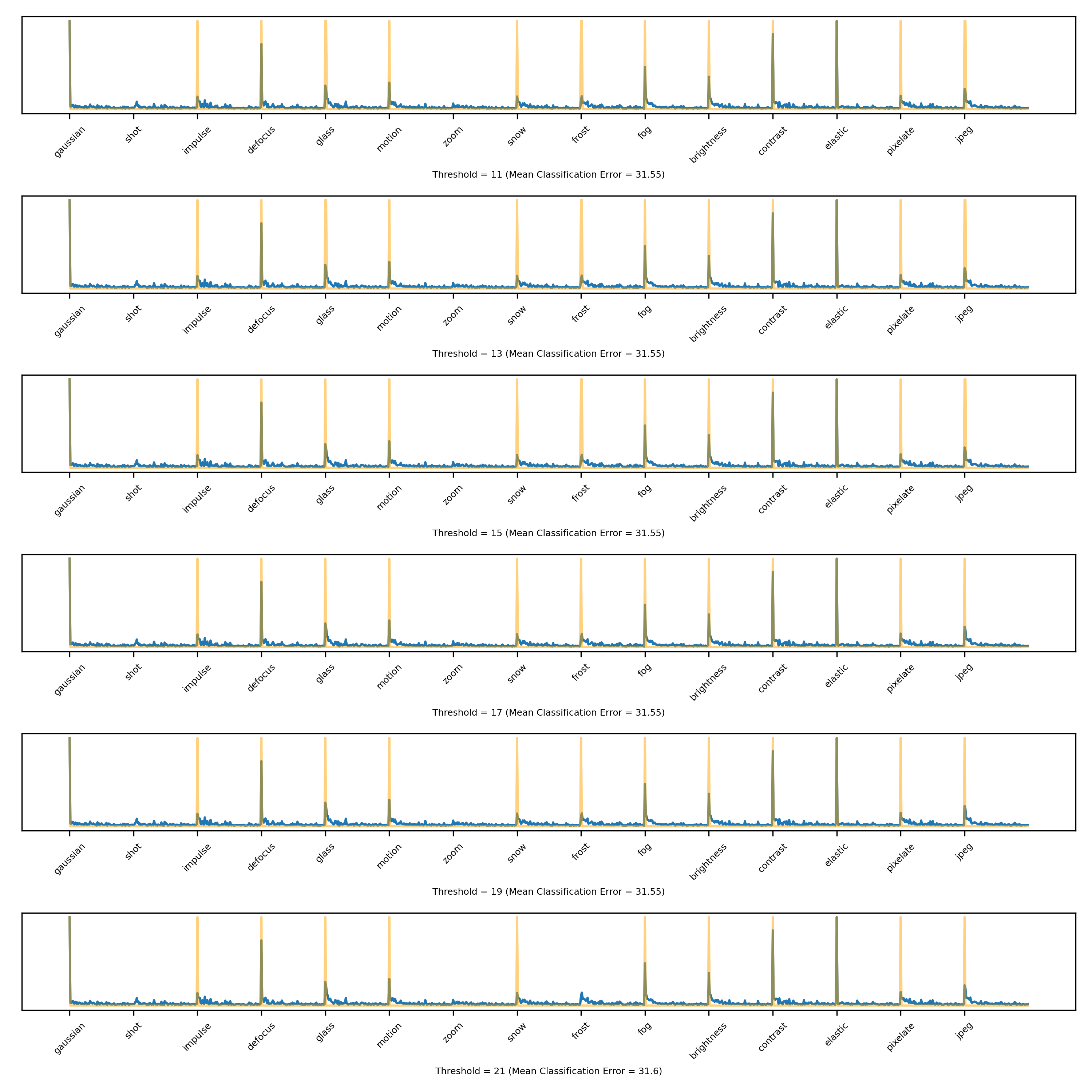} 
\vspace{-20pt}
\caption{Decision diagram corresponding to different threshold values of online peak detection algorithm on CIFAR-100C. It can be observed that our method is less sensitive to different values of the threshold and yields same classification performance.}
\label{fig:ablation}
\end{figure*}

\begin{figure*}[t!]
\centering
\includegraphics[width=1\textwidth]{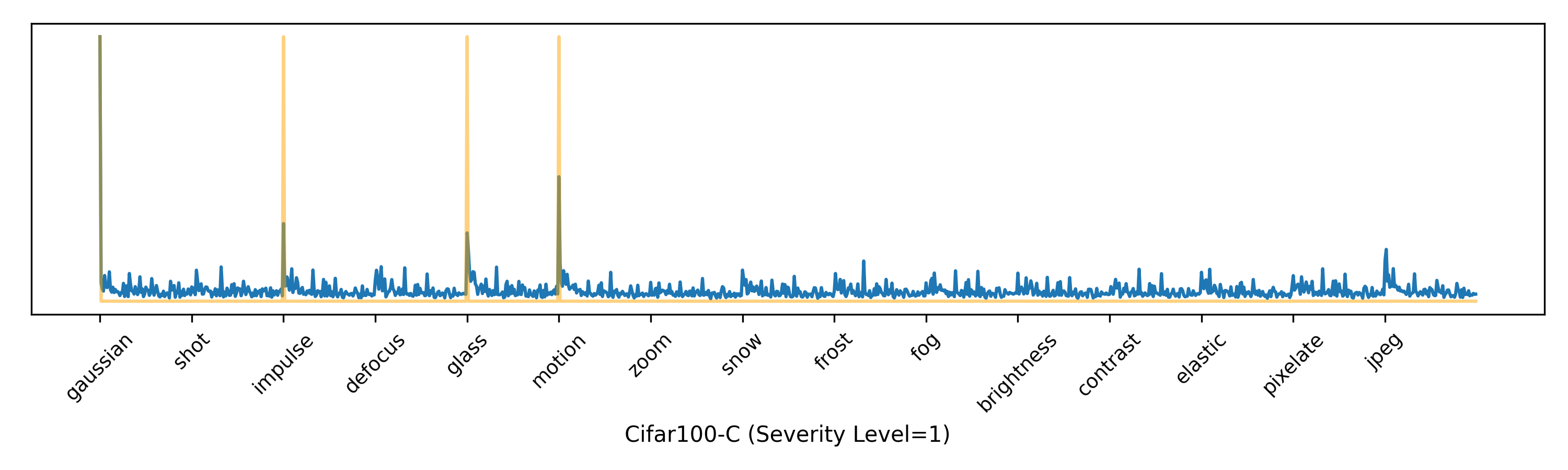} 
\vspace{-20pt}
\caption{Decision diagram corresponding to CIFAR100-C with severity level 1.}
\label{fig:limitation}
\end{figure*}

From Fig. \ref{fig:ablation}, it can be observed that our method is less sensitive to different values of the threshold and yields same classification performance. However, excessively high values could potentially overlook peak values, as observed by the last subplot of Fig. \ref{fig:ablation}.


\section{Limitations}
If two domains share very similar features, i.e., domain gap is very low, our method's ability to detect domain changes can be diminished, as the global BN statistics between the two domains will be similar. However, it should also be noted that when the domain gap is low, the problem of forgetting and error accumulation are very much reduced, hence the need for taking additional measure to address the problems remains less critical. To demonstrate this, we perform the same classification experiment as Table 1 of main paper on CIFAR-100C, but with a severity level 1. Such low severity level results in images of different domains to be almost indistinguishable to the human eye. The results are shown in Table \ref{tab:limitation}.

From the table it can be observed that Tent-Continual exhibits only a marginal performance decline of $2.3\%$ compared to Tent-Online. This is in contrast to the experiment in Table 1 of the main paper with a severity level of 5, which led to a performance drop of $37.1\%$. Clearly, in scenarios with minimal domain gap, the challenges of error accumulation and forgetting are significantly mitigated. Furthermore, it's worth highlighting that even in such a challenging scenario of low domain gap, our method has managed to enhance performance when integrated with Tent. 

Additionally, in Figure \ref{fig:limitation}, we have visualized the decision thresholds for the severity 1 case. The figure highlights that the absence of a peak in BN statistics is noticeable when encountering new domains, primarily due to the similarity of global BN statistics across domains. Despite this challenging situation, our method successfully identifies several domain changes, leading to performance enhancement over the continual model, as demonstrated in Table \ref{tab:limitation}.

\null
\vfill
